%% file: main.tex
\documentclass[sigconf]{acmart}

\makeatletter
\DeclareRobustCommand*\cal{\@fontswitch\relax\mathcal}
\makeatother

\usepackage{booktabs} 
\usepackage{amsmath,amsfonts}
\usepackage{graphicx}
\usepackage{subfig}
\usepackage{textcomp}
\usepackage{balance}
\usepackage{xcolor}
\usepackage{caption}
\usepackage{geometry}
\usepackage[export]{adjustbox}
\usepackage{tabularx}
\usepackage{graphicx}
\usepackage{latexsym}
\usepackage{hyperref}
\usepackage{bm}
\usepackage{algpseudocode}
\usepackage{algorithm}

\algdef{SE}[SUBALG]{Indent}{EndIndent}{}{\algorithmicend\ }%
\algtext*{Indent}
\algtext*{EndIndent}

\setcopyright{none}

\copyrightyear{2020}
\acmYear{2020}

\acmConference[GECCO '20 Companion]{Genetic and Evolutionary Computation Conference Companion}{July 8--12, 2020}{Canc\'{u}n, Mexico}

\acmPrice{15.00}
\acmDOI{10.1145/3377929.3398123}
\acmISBN{978-1-4503-7127-8/20/07}

\begin{document}
\title[A Modular Hybridization of PSO and DE]{A Modular Hybridization of Particle Swarm Optimization and Differential Evolution}

\author{Rick Boks}
\affiliation{%
  \institution{Leiden Institute of Advanced Computer Science}
  \city{Leiden} 
  \country{The Netherlands} 
}
\email{r.m.boks@umail.leidenuniv.nl}

\author{Hao Wang}
\affiliation{
  \institution{LIP6, Sorbonne Universit\'e}
  \city{Paris} 
  \country{France} 
}
\email{hao.wang@lip6.fr}

\author{Thomas B\"ack}
\affiliation{%
  \institution{Leiden Institute of Advanced Computer Science}
  \city{Leiden} 
  \country{The Netherlands} 
}
\email{t.h.w.baeck@liacs.leidenuniv.nl}

\renewcommand{\shortauthors}{R. Boks et al.}

\begin{abstract}
In swarm intelligence, Particle Swarm Optimization (PSO) and Differential Evolution (DE) have been successfully applied in many optimization tasks, and a large number of variants, where novel algorithm operators or components are implemented, has been introduced to boost the empirical performance.
In this paper, we first propose to combine the variants of PSO or DE by modularizing each algorithm and incorporating the variants thereof as different options of the corresponding modules. Then, considering the similarity between the inner workings of PSO and DE, we hybridize the algorithms by creating two populations with variation operators of PSO and DE respectively, and selecting individuals from those two populations. The resulting novel hybridization, called PSODE, encompasses most up-to-date variants from both sides, and more importantly gives rise to an enormous number of unseen swarm algorithms via different instantiations of the modules therein. 

In detail, we consider $16$ different variation operators originating from existing PSO- and DE algorithms, which, combined with $4$ different selection operators, allow the hybridization framework to generate $800$ novel algorithms. The resulting set of hybrid algorithms, along with the combined $30$ PSO- and DE algorithms that can be generated with the considered operators, is tested on the $24$ problems from the well-known COCO/BBOB benchmark suite, across multiple function groups and dimensionalities. 
\end{abstract}

\maketitle

\input{body}

\section*{Acknowledgments} 
Hao Wang acknowledges the support from the Paris \^Ile-de-France Region.

\bibliographystyle{ACM-Reference-Format}
\bibliography{bibliography} 
\end{document}

%% file: body.tex
\section{Introduction}
In this paper, we delve into two naturally-inspired algorithms, Particle Swarm Optimization (PSO)~\cite{rjnewoptimizer} and Differential Evolution (DE)~\cite{firstde} for solving continuous black-box optimization problems $f:\mathbb{R}^n \rightarrow \mathbb{R}$, which is subject to minimization without loss of generality. Here we only consider simple box constraints on $\mathbb{R}^n$, meaning the search space is a hyper-box $[\mathbf{x}^{\text{min}}, \mathbf{x}^{\text{max}}] = \prod_{i=1}^n [x^{\text{min}}_i, x^{\text{max}}_i]$.

In the literature, a huge number of variants of PSO and DE has been proposed to enhance the empirical performance of the respective algorithms. Despite the empirical success of those variants, we, however, found that most of them only differ from the original PSO/DE in one or two operators (e.g., the crossover), where usually some simple modifications are implemented. Therefore, it is almost natural for us to consider combinations of those variants. Following the so-called configurable CMA-ES approach~\cite{es,vanRijn:2017:ACD:3071178.3071205}, we first modularize both PSO and DE algorithms, resulting in a modular framework where different types of algorithmic modules are applied sequentially in each generation loop. When incorporating variants into this modular framework\footnote{The source code is available at \url{https://github.com/rickboks/pso-de-framework}.}, we first identify the modules at which modifications are made in a particular variant, and then treat the modifications as options of the corresponding modules. For instance, the so-called inertia weight~\cite{inertiaweight}, that is a simple modification to the velocity update in PSO, shall be considered as an option of the velocity update module.

This treatment allows for combining existing variants of either PSO or DE and generating non-existing algorithmic structures. It, in the loose sense, creates a \emph{space/family of swarm algorithms}, which is configurable via instantiating the modules, and hence potentially primes the application of algorithm selection/configuration~\cite{10.1145/2487575.2487629} to swarm intelligence. More importantly, we also propose a meta-algorithm called \textbf{PSODE} that \emph{hybridizes} the variation operators from both PSO and DE, and therefore gives rise to an even larger space of unseen algorithms. 
By hybridizing PSO and DE, we aim to unify the strengths from both sides, in an attempt to, for instance, improve the population diversity and the convergence rate.
On the well-known Black-Box Optimization Benchmark (BBOB)~\cite{coco} problem set, we extensively tested all combinations of four different velocity updates (PSO), five neighborhood topologies (PSO), two crossover operators (DE), five mutation operators (DE), and four selection operators, leading up to $800$ algorithms. We benchmark those algorithms on all $24$ test functions from the BBOB problem set and analyze the experimental results using the so-called IOHprofiler~\cite{doerr2019benchmarking}, to identify algorithms that perform well on (a subset of) the 24 test functions.
 
This paper is organized as follows: Section~\ref{ch:relatedwork} summarizes the related work. Section~\ref{ch:pso} reviews the state-of-the-art variants of PSO. Section~\ref{ch:de} covers various cutting-edge variants of DE. In Section~\ref{sec:hybrid}, we describe the novel modular PSODE algorithm. Section~\ref{ch:experiment} specifies the experimental setup on the BBOB problem set.
We discuss the experimental results in Section~\ref{ch:results} and finally provide, in Section~\ref{ch:conclusion}, the insights obtained in this paper as well as future directions.

\section{Related Work} \label{ch:relatedwork}
A hybrid PSO/DE algorithm has been coined previously~\cite{depso} to improve the population diversity and prevent premature convergence. This is attempted by using the DE mutation instead of the traditional velocity- and position-update to evolve candidate solutions in the PSO algorithm. This mutation is applied to the particle's best-found solution $\mathbf{p}_i$ rather than its current position $\mathbf{x}_i$, resulting in a steady-state strategy. Another approach~\cite{hendtlass} follows the conventional PSO algorithm, but occasionally applies the DE operator in order to escape local minima. Particles maintain their velocity after being permuted by the DE operator. Other PSO/DE hybrids include a two-phase approach~\cite{twophasehybrid} and a Bare-Bones PSO variant based on DE~\cite{debasedpso}, which requires little parameter tuning.

This work follows the approach of the modular and extensible CMA-ES framework proposed in \cite{es}, where many \emph{ES-structures} can be instantiated by arbitrarily combining existing variations of the CMA-ES. The authors of this work implement a Genetic Algorithm to efficiently evolve the ES structures, instead of performing an expensive brute force search over all possible combinations of operators.

\section{Particle Swarm Optimization} \label{ch:pso}
As introduced by Eberhart and Kennedy~\cite{rjnewoptimizer}, Particle Swarm Optimization (PSO) is an optimization algorithm that mimics the behaviour of a flock of birds foraging for food. A particle in a swarm of size $M \in \mathbb{N}_{>1}$ is associated with three vectors: the current position $\mathbf{x}_i$, 
velocity $\mathbf{v}_i$, and its previous best position $\mathbf{p}_i$, where $i \in \{1,\ldots,M\}$.
After the initialization of $\mathbf{x}_i$ and $\mathbf{v}_i$, where $\mathbf{x}_i$ is initialized randomly and $\mathbf{v}_i$ is set to $\mathbf{0}$, the algorithm iteratively controls the velocity $\mathbf{v}_i$ for each particle (please see the next subsection) and moves the particle $\mathbf{x}_i$ accordingly:
\begin{equation}
\mathbf{x}_{i} \leftarrow \mathbf{x}_{i} + \mathbf{v}_{i}
\end{equation}
To prevent the velocity from exploding,
$\mathbf{v}_i$ is kept in the range $[-v_{\text{max}}\mathbf{1}, v_{\text{max}}\mathbf{1}]$ ($\mathbf{1}$ is a $n\times 1$ vector containing all ones).
After every position update, the current position is evaluated, $f_i = f(\mathbf{x}_i)$. Here, $\mathbf{p}_i$ stands for the best solution found by $\mathbf{x}_i$ (thus personal best) while $\mathbf{g}_i$ is used to track the best solution found in the \emph{neighborhood} of $\mathbf{x}_i$ (thus global best). Typically, the termination of PSO can be determined by simple termination criteria, such as the depletion of the function evaluation budget, as well as more complicated ones that reply on the convergence behavior, e.g., detecting whether the average distance between particles has gone below a predetermined threshold. The pseudo-code is given in Alg.~\ref{Alg:PSO}.

\begin{algorithm}[!htbp]
\begin{algorithmic}[1]
	\For{$i = 1 \rightarrow M$}
		\State{$f^{\text{best}}_i \leftarrow f(\mathbf{x}_i)$}
		\State{$\mathbf{x}_i \leftarrow \bm{U}(\mathbf{x}^{\text{min}}, \mathbf{x}^{\text{max}}), \quad \mathbf{v}_i \leftarrow \mathbf{0}$} \Comment{Initialize} 
	\EndFor
	
	\While{termination criteria are not met}
		\For{$i = 1 \rightarrow M$}
			\State{$f_i \leftarrow f(\mathbf{x}_i)$}\Comment{Evaluate}
			\If{$f_i < f^{\text{best}}_i$}
				\State{$\mathbf{p}_i \leftarrow \mathbf{x}_i, \quad f^{\text{best}}_i \leftarrow f_i$}\Comment{Update personal best}
			\EndIf   
			\If{$f_i < f(\mathbf{g}_i)$}
				\State{$\mathbf{g}_i \leftarrow \mathbf{x}_i$} \Comment{Update global best}
			\EndIf  
			\State{Calculate $\mathbf{v}_i$ according to Eq.~\eqref{eq:velocity-original}}
			\State{$\mathbf{x}_{i} \leftarrow \mathbf{x}_{i} + \mathbf{v}_{i}$}  \Comment{Update position} 
		\EndFor
	\EndWhile
 \end{algorithmic}
\caption{Original Particle Swarm Optimization}
\label{Alg:PSO}
\end{algorithm}
\vspace{-2mm}

\subsection{Velocity Updating Strategies} \label{sec:velocityupdate}
As proposed in the original paper~\cite{rjnewoptimizer}, the velocity vector in \emph{original} PSO is updated as follows:
\begin{equation}
\label{eq:velocity-original}
\mathbf{v}_{i} \leftarrow \mathbf{v}_{i} + \bm{U}(\mathbf{0},\phi_{1}\mathbf{1}) \otimes (\mathbf{p}_i - \mathbf{x}_{i}) + \bm{U}(\mathbf{0},\phi_{2}\mathbf{1}) \otimes (\mathbf{g}_i - \mathbf{x}_i),
\end{equation}
where $\bm{U}(\mathbf{a}, \mathbf{b})$ stands for a continuous uniform random vector with each component distributed uniformly in the range $[a_i, b_i]$, and $\otimes$ is component-wise multiplication. Note that, henceforth the parameter settings such as $\phi_1, \phi_2$ will be specified in the experimentation part (Section~\ref{ch:experiment}). As discussed before, velocities resulting from Eq.~\eqref{eq:velocity-original} have to be clamped in range $[-v_{\text{max}}\mathbf{1}, v_{\text{max}}\mathbf{1}]$. Alternatively, the \emph{inertia weight}~\cite{inertiaweight} $\omega\in [0, 1]$ is introduced to moderate the velocity update without using $v_{\text{max}}$:
\begin{equation}
\label{eq:velocity-inertia}
\mathbf{v}_{i} \leftarrow \omega\mathbf{v}_{i} + \bm{U}(\mathbf{0},\phi_{1}\mathbf{1}) \! \otimes \! (\mathbf{p}_i - \mathbf{x}_{i}) + \bm{U}(\mathbf{0},\phi_{2}\mathbf{1}) \! \otimes \! (\mathbf{g}_i - \mathbf{x}_i).
\end{equation}
A large value of $\omega$ will result in an exploratory search, while a small value leads to a more exploitative behavior. It is suggested to decrease the inertia weight over time as it is desirable to scale down the explorative effect gradually. Here, we consider the inertia method with fixed as well as decreasing weights. 
\par
Instead of only being influenced by the best neighbor, the velocity of a particle in the \emph{Fully Informed Particle Swarm} (FIPS)~\cite{fips} is updated using the best previous positions of \textit{all} its neighbors. The corresponding equation is:
\begin{equation}
\label{eq:velocity-FIPS}
    \mathbf{v}_i \leftarrow \chi\Big(\mathbf{v}_i + \frac{1}{|N_i|} \sum_{\mathbf{p} \in N_i}\bm{U}(\mathbf{0},\phi\mathbf{1}) \otimes (\mathbf{p} - \mathbf{x}_i) \Big),
\end{equation}
where $N_i$ is the number of neighbors of particle $i$ and $\chi= 2/(\phi - 2 + \sqrt{\phi^2 -4\phi})$. Finally, the so-called \emph{Bare-Bones} PSO~\cite{barebones} is a completely different approach in the sense that velocities are not used at all and instead every component $x_{ij}$ ($j=1,\ldots,n$) of position $\mathbf{x}_i$ is sampled from a Gaussian distribution with mean $(p_{ij} + g_{ij})/2$ and variance $|p_{ij} - g_{ij}|$, where $p_{ij}$ and $g_{ij}$ are the $j$th component of $\mathbf{p}_i$ and $\mathbf{g}_i$, respectively:
\begin{equation}\label{eq:barebones}
x_{ij} \sim {\cal N}\left((p_{ij}+g_{ij})/2, |p_{ij}-g_{ij}|\right), \quad j=1,\ldots, n.
\end{equation}

\subsection{Population Topologies} \label{sec:topologies}
Five different topologies from the literature have been implemented in the framework:
\begin{itemize}
	\item \textit{lbest} (local best)~\cite{rjnewoptimizer} takes a ring topology and each particle is only influenced by its two adjacent neighbors.
	\item \textit{gbest} (global best)~\cite{rjnewoptimizer} uses a fully connected graph and thus every particle is influenced by the best particle of the entire swarm.
	\item In the \textit{Von Neumann topology}~\cite{popstructure}, particles are arranged in a two-dimensional array and have four neighbors: the ones horizontally and vertically adjacent to them, with toroidal wrapping.
	\item The \emph{increasing topology}~\cite{neighborhoodoperator} starts with an \textit{lbest} topology and gradually increases the connectivity so that, by the end of the run, the particles are fully connected.
	\item The \emph{dynamic multi-swarm topology} (DMS-PSO)~\cite{multiswarm} creates clusters consisting of three particles each, and creates new clusters randomly after every $5$ iterations. If the population size is not divisible by three, every cluster has size three, except one, which is of size $3 + (M \!\mod 3)$.
\end{itemize}

\section{Differential Evolution}
\label{ch:de}
Differential Evolution (DE) is introduced by Storn and Price in 1995~\cite{firstde} and 
uses scaled differential vectors between randomly selected individuals for perturbing the population. The pseudo-code of DE is provided in Alg.~\ref{alg:de}.

After the initialization of the population (please see the next subsection) $P=\{\mathbf{x}_i\}_{i=1}^M \subset \mathbb{R}^n$ ($M$ is again the swarm size), for each individual $\mathbf{x}_i$, a donor vector $\mathbf{v}_i$ (a.k.a.~mutant) is generated according to:
\begin{equation} \label{eq:DE/rand/1}
\mathbf{v}_i \leftarrow \mathbf{x}_{r_1} + F \cdot (\mathbf{x}_{r_2} - \mathbf{x}_{r_3})
\end{equation}
where three distinct indices $r_1 \neq r_2 \neq r_3 \neq i \in [1..M]$ are chosen uniformly at random (u.a.r.). Here $F \in [0.4 , 1]$ is a scalar value called the \textit{mutation rate} and $\mathbf{x}_{r_1}$ is referred as the \emph{base vector}. Afterwards, a \textit{trial vector} $\mathbf{x}_i'$ is created by means of crossover.

In the so-called binomial crossover, each component $x_{ij}'$ ($j=1,\ldots, n$) of $x_i'$ is copied from $v_{ij}$ with a probability $Cr\in[0, 1]$ (a.k.a.~crossover rate), or when $j$ equals an index $j_{\text{rand}} \in [1..n]$ chosen u.a.r.:
\begin{equation}
\label{eq:binomial}
x_{ij}' \leftarrow \left\{
  \begin{array}{l}
    v_{ij} \quad\text{ if } U(0,1) \leq Cr \text{ or } j = j_{\text{rand}} \\
    x_{ij} \quad \text { otherwise}
  \end{array}
\right.
\end{equation}

In exponential crossover, two integers $p$, $q$  $\in \{1, \dots,n\}$ are chosen. The integer $p$ acts as the starting point where the exchange of components begins, and is chosen uniformly at random. $q$ represents the number of elements that will be inherited from the donor vector, and is chosen using Algorithm~\ref{alg:setq}.
\begin{algorithm}
\begin{algorithmic}[1]
\State{$q \leftarrow 0$}

\State{\textbf{do}}
\Indent
\State{$q \leftarrow q+1$}
\EndIndent
\State{\textbf{while} $((U(0,1) \leq Cr)$ and $(q \leq n))$}

\end{algorithmic}
\caption{Assigning a value to $q$}
\label{alg:setq}
\end{algorithm}

The trial vector $\mathbf{x}'_i$ is generated as:
\begin{equation}
\label{eq:exponential}
\hspace{-0.43em} x'_{ij} \leftarrow \! \left\{
  \begin{array}{l}
    v_{ij} \text{ for } j = \langle p \rangle_n, \langle p +1 \rangle_n \dots \langle p + q - 1 \rangle_n\\
    x_{ij} \text { for all other } j \in \{1, \dots,n\}\\
  \end{array}
\right.
\end{equation}

The angular brackets $\langle \rangle_n$ denote the modulo operator with modulus $n$. Elitism selection is applied between $\mathbf{x}_i$ and $\mathbf{x}_i'$, where the better one is kept for the next iteration.

\begin{algorithm}[!ht]
\begin{algorithmic}[1]
	\State{$\mathbf{x}_i \leftarrow \bm{U}(\mathbf{x}^{\text{min}}, \mathbf{x}^{\text{max}}), \quad i =1, \ldots, M.$}\Comment{Initialize}
	\While{termination criteria are not met}
		\For{$i = 1 \rightarrow M$}
			\State{Choose $r_1 \neq r_2 \neq r_3 \neq i \in [1..M]$ u.a.r.}
			\State{$\mathbf{v}_i \leftarrow \mathbf{x}_{r_{1}} + F(\mathbf{x}_{r_{2}} - \mathbf{x}_{r_{3}})$}\Comment{Mutate}
			\State{Choose $j_{\text{rand}} \in [1..n]$ u.a.r.}
			\For {$j = 1 \rightarrow n$}
    			\If{$U(0,1) \leq Cr$ or $j = j_{\text{rand}}$}
    				\State $x_{ij}' \leftarrow v_{ij}$
                \Else
                    \State $x_{ij}' \leftarrow x_{ij}$
                \EndIf
            \EndFor
            \If{$f(\mathbf{x}_i') < f(\mathbf{x}_i)$}
				\State $\mathbf{x}_i \leftarrow \mathbf{x}_i'$ \Comment{Select}	
			\EndIf
		\EndFor
	\EndWhile
\caption{Differential Evolution using Binomial Crossover}
\label{alg:de}
\end{algorithmic}
\end{algorithm}

\subsection{Mutation} 
\label{sec:mutation}
In addition to the so-called DE/rand/1 mutation operator (Eq.~\ref{eq:DE/rand/1}), we also consider the following variants:
\begin{enumerate}
    \item DE/best/1~\cite{firstde}: the base vector is chosen as the current best solution in the population $\mathbf{x}_{\text{best}}$:
    $$\mathbf{v}_i \leftarrow \mathbf{x}_{\text{best}} + F \cdot (\mathbf{x}_{r_{1}} - \mathbf{x}_{r_{2}})$$
    \item DE/best/2~\cite{firstde}: two differential vectors calculated using four distinct solutions are scaled and combined with the current best solution:
   	$$\mathbf{v}_i \leftarrow \mathbf{x}_{\text{best}} + F \cdot (\mathbf{x}_{r_{1}} - \mathbf{x}_{r_{2}}) + F \cdot (\mathbf{x}_{r_{3}} - \mathbf{x}_{r_{4}})$$
    \item DE/Target-to-best/1~\cite{firstde}: the base vector is chosen as the solution on which the mutation will be applied and the difference from the current best to this solution is used as one of the differential vectors:
    $$\mathbf{v}_i \leftarrow \mathbf{x}_{i} + F \cdot (\mathbf{x}_{\text{best}} - \mathbf{x}_{i}) + F \cdot (\mathbf{x}_{r_{1}} - \mathbf{x}_{r_{2}})$$
    \item Target-to-$p$best/1~\cite{jade}: the same as above except that we take instead of the current best a solution $\textbf{x}_{\text{best}}^p$ that is randomly chosen from the top 100$p\%$ solutions in the population with $p \in (0,1]$.
    $$\mathbf{v}_i \leftarrow \mathbf{x}_{i} + F \cdot (\mathbf{x}_{\text{best}}^p - \mathbf{x}_{i}) + F \cdot (\mathbf{x}_{r_{1}} - \mathbf{x}_{r_{2}})$$
    \item DE/2-Opt/1~\cite{twoopt1}: 
    $$\mathbf{v}_i \leftarrow \left\{
  \begin{array}{l}
   \mathbf{v}_i \leftarrow \mathbf{x}_{r_1} + F(\mathbf{x}_{r_2} - \mathbf{x}_{r_3}) \quad\text{ if } f(\mathbf{x}_{r_1}) < f(\mathbf{x}_{r_2}) \\
    \mathbf{v}_i \leftarrow \mathbf{x}_{r_2} + F(\mathbf{x}_{r_1} - \mathbf{x}_{r_3}) \quad\text{ otherwise}
  \end{array}
  \right.$$
\end{enumerate}
\noindent

\subsection{Self-Adaptation of Control Parameters}
The performance of the DE algorithm is highly dependent on values of the parameters $F$ and $Cr$, for which the optimal values are in turn dependent on the optimization problem at hand. The self-adaptive DE variant JADE~\cite{jade} has been proposed in desire to control the parameters in a self-adaptive manner, without intervention of the user. This self-adaptive parameter scheme is used in both DE and hybrid algorithm instances.

\section{Hybridizing PSO with DE} \label{sec:hybrid}
Here, we propose a hybrid algorithm framework called \textbf{PSODE}, that combines the mutation- and crossover operators from DE with the velocity- and position updates from PSO. This implementation allows combinations of all operators mentioned earlier, in a single algorithm, creating the potential for a large number of possible hybrid algorithms. We list the pseudo-code of PSODE in Alg.~\ref{Alg:PSODE}, which works as follows. 

\clearpage
\begin{enumerate}
	\item The initial population $P_0 = \{\mathbf{x}^{(1)}, \ldots, \mathbf{x}^{(M)}\}$ ($M$ stands for the swarm size) is sampled uniformly at random in the search space, and the corresponding velocity vectors are initialized to zero (as suggested in~\cite{6256112}).
	\item After evaluating $P_0$, we create $P_1$ by applying the PSO position update to each solution in $P_0$. 
	\item Similarly, $P_2$ is created by applying the DE mutation to each solution in $P_0$.
	\item Then, a population $P_3$ of size $M$ is generated by recombining information among the solutions in $P_0$ and $P_2$, based on the DE crossover. 
	\item Finally, a new population is generated by selecting good solutions from $P_0, P_1,$ and $P_3$ (please see below).
	\end{enumerate}  
Four different selection methods are considered in this work, two of which are elitist, and two non-elitist. A problem arises during the selection procedure: solutions from $P_3$ have undergone the mutation and crossover of DE that alters their positions but ignores the velocity thereof, leading to an unmatched pair of positions and velocities. In this case, the velocities that these particles have inherited from $P_0$ may no longer be meaningful, potentially breaking down
the inner workings of PSO in the next iteration. To solve this issue, we propose to re-compute the velocity vector according to the displacement of a particle resulting from mutation and crossover operators, namely:
\begin{equation}\label{eq:velocity-recaculate}
\mathbf{v}^{(i)} \leftarrow \mathbf{x}^{(i, 3)} - \mathbf{x}^{(i, 0)}, \text{ for } i=1,2,\ldots, M,
\end{equation}
where $\mathbf{x}^{(i, 3)} \in P_3$ is generated by $\mathbf{x}^{(i, 0)} \in P_0$ using aforementioned procedure. 

A selection operator is required to select particles from $P_0$, $P_1$, and $P_3$ for the next generation. Note that $P_2$ is not considered in the selection procedure, as the solution vectors in this population were recombined and stored in $P_3$. We have implemented four different selection methods: two of those methods only consider population $P_1$, resulting from variation operators of PSO, and population $P_3$, obtained from variation operators of DE. This type of selection methods is essentially non-elitist allowing for deteriorations. Alternatively, the other two methods implement elitism by additionally taking population $P_0$ into account. 

We use the following naming scheme for the selection methods:
$$\texttt{[comparison method]/}\texttt{[\#} P_i \texttt{ considered]}$$

Using this scheme, we can distinguish the four selection methods:
pairwise/2, pairwise/3, union/2, and union/3. The ``pairwise'' comparison
method means that the $i$-th members (assuming the solutions are indexed) of each considered population are compared to each other, from which we choose the best one for the next generation. The ``union'' method selects the best $M$ solutions from the union of the considered populations. Here, a ``2'' signals the inclusion of two populations, $P_1$ and $P_3$, and a ``3'' indicates the further inclusion of $P_0$. For example, the pairwise/2 method selects the best individual from each pair of $\textbf{x}^{(i, 1)}$ and $\textbf{x}^{(i, 3)}$, while the union/3 method selects the best $M$ individuals from $P_0 \cup P_1 \cup P_3$.

\begin{algorithm}[!htbp]
\begin{algorithmic}[1]
	\State{Sample $P_0= \{\mathbf{x}^{(1)}, \ldots, \mathbf{x}^{(M)}\}$ uniformly at random in $[\mathbf{x}^{\text{min}},\mathbf{x}^{\text{max}}]$}
	\State{Initialize velocities $V \leftarrow \{\mathbf{0}, \ldots, \mathbf{0}\}$.}
	\While{termination criteria are not met}
		\State $P_1\leftarrow \emptyset$
		\For{$\mathbf{x}\in P_0$ with its corresponding velocity $\mathbf{v}\in V$}
			\State $\mathbf{v}' \leftarrow \textsc{velocity-update}(\mathbf{x}, \mathbf{v})$
			\State $\mathbf{x}' \leftarrow \mathbf{x} + \mathbf{v}'$
			\State Evaluate $\mathbf{x}'$ on $f$
			\State $P_1 \leftarrow P_1 \cup \{\mathbf{x}'\}$
		\EndFor
		\State $P_2\leftarrow \emptyset$
		\For{$\mathbf{x} \in P_0$}
			\State $\mathbf{x}' \leftarrow \textsc{de-mutation}(\mathbf{x})$
			\State $P_2 \leftarrow P_2 \cup \{\mathbf{x}'\}$
		\EndFor
		\State $P_3\leftarrow \emptyset, V\leftarrow\emptyset$
		\For{$i = 1 \to M$}
			\State $\mathbf{x}' \leftarrow \textsc{de-crossover}(\mathbf{x}^{(i,0)}, \mathbf{x}^{(i,2)})$
			\State calculate $\mathbf{v}'$ for $\mathbf{x}'$ using Eq.~\ref{eq:velocity-recaculate}
			\State Evaluate $\mathbf{x}'$ on $f$
			\State $P_3 \leftarrow P_3 \cup \{\mathbf{x}'\}$
			\State $V \leftarrow V \cup \{\mathbf{v}'\}$
		\EndFor
		\State $P_0 \leftarrow \textsc{selection}(P_0, P_1, P_3)$
	\EndWhile
 \end{algorithmic}
\caption{PSODE}
\label{Alg:PSODE}
\end{algorithm}

\vfill\null
\section{Experiment} \label{ch:experiment}
A software framework has been implemented  in \texttt{C++} to generate PSO, DE and PSODE instances from all aforementioned algorithmic modules, e.g. topologies and mutation strategies. Such a framework is tested on IOHprofiler, which contains the $24$ functions from BBOB/COCO~\cite{coco} that are organized in five function groups: 1) Separable functions 2) Functions with low or moderate conditioning 3) Unimodal functions with high conditioning 4) Multi-modal functions with adequate global structure and 5) Multi-modal functions with weak global structure. 

In the experiments conducted, a PSODE instance is considered as a combination of five modules: velocity update strategy, population topology, mutation method, crossover method, and selection method. Combining each option for each of these five modules, we obtain a total of $5 \text{ ( topologies )} \times 4 \text{ ( velocity strategies )} \times 5 \text{ ( mutation methods ) } \times 2 \text{ ( crossover methods ) } \times 4 \text { ( selection } \\ \text{methods )} = 800$ different PSODE instances.

By combining the $4$ velocity update strategies and $5$ topologies, we obtain $4 \times 5 = 20$ PSO instances, and similarly we obtain $5 \text{ ( mutation methods ) } \times 2 \text { ( crossover methods ) } = 10$ DE instances.

\clearpage
\paragraph{Naming Convention of Algorithm Instances}
 As each PSO, DE, and hybrid instance can be specified by the composing modules, it is named using the abbreviations of its modules: hybrid instances are named as follows: \\
 
\noindent
\begin{center}
\texttt{H\_[velocity strategy]\_[topology]\_[mutation] \\
\_[crossover]\_[selection]} \\
\end{center}

\noindent
PSO instances are named as:
$$\texttt{P\_[velocity strategy]\_[topology]}$$
And DE instances are named as: 
$$\texttt{D\_[mutation]\_[crossover]}$$
Options of all modules are listed in Table~\ref{tab:naming}.

\paragraph{Experiment Setup} The following parameters are used throughout the experiment:
\begin{itemize}
	\item Function evaluation budget: $10^4n$.
	\item Population (swarm) size: $5n$ is used for all algorithm instances, due to the relatively consistent performance that instances show across different function groups and dimensionalities when using this value.
	\item Hyperparameters in PSO: In Eq.~\eqref{eq:velocity-original} and~\eqref{eq:velocity-inertia}, $\phi_1 = \phi_2 = 1.49618$ is taken as recommended in~\cite{constrictioncoeff} and for FIPS (Eq.~\eqref{eq:velocity-FIPS}), a setting $\phi=4.1$ is adopted from~\cite{fips}. In the fixed inertia strategy, $\omega$ is set to $0.7298$ while in the decreasing inertia strategy, $\omega$ is linearly decreased from $0.9$ to $0.4$. For the Target-to-$p$best/1 mutation scheme, a value of $p = 0.1$ is chosen, following the findings of~\cite{jade}.
	\item Hyperparameters in DE: $F$ and $Cr$ are managed by the JADE self-adaptation scheme.
	\item Number of independent runs per function: $30$. Note that only one \emph{function instance} (instance ``1'') is used for each function.
	\item Performance measure: 
	\emph{expected running time} (ERT)~\cite{ert}, which is the total number of function evaluations an algorithm is expected to use to reach a given target function value for the first time. ERT is defined as $\frac{\#\text{FEs}(f_{\text{target}})}{\#\text{succ}}$, where $\#\text{FEs}(f_{\text{target}})$ denotes the total number of function evaluations taken to hit $f_{\text{target}}$ in all runs, while $f_{\text{target}}$ might not be reached in every run, and $\#\text{succ}$ denotes the number of successful runs. 

\end{itemize} 
To present the result, we rank the $830$ algorithm instances with regard to their ERT values. This is done by first ranking the instances on the targets $f_{\text{opt}} + \{10^1, \dots ,10^{-8}\}$ of every benchmark function, and then taking the \textit{average rank} across all targets per function. Finally, the presented rank is obtained by taking the average rank over all $24$ test functions. This is done for both dimensionalities. A dataset containing the running time for each independent run and ERT's for each algorithm instance, with the supporting scripts, are available at~\cite{rick_boks_2020_3814197}.

\begin{small}
\begin{table}[H]
\begin{tabular}{l|l}
\multicolumn{1}{c}{\texttt{[velocity strategy]}} \vline & \multicolumn{1}{c}{\texttt{[mutation]}} \\
\hline
\texttt{B} --  Bare-Bones PSO &  \texttt{B1} -- DE/best/1\\
\texttt{F} -- Fully-informed PSO (FIPS) &  \texttt{B2} -- DE/best/2  \\
\texttt{I} -- Inertia weight &  \texttt{T1} -- DE/target-to-best/1\\
\texttt{D} -- Decreasing inertia weight & \texttt{PB} -- DE/target-to-$p$best/1\\
\cline{1-1} 
\multicolumn{1}{c}{\texttt{[crossover]}}\vline  &\texttt{O1} -- 2-Opt/1 \\ \cline{1-2} 
\texttt{B} -- Binomial crossover&\multicolumn{1}{c}{\texttt{[selection]}}  \\
\cline{2-2} 
\texttt{E} -- Exponential crossover& \texttt{U2} -- Union/2 \\ \cline{1-1} 
\multicolumn{1}{c}{\texttt{[topology]}} \vline& \texttt{U3} -- Union/3 \\ \cline{1-1} 
\texttt{L} -- $lbest$ (ring)&  \texttt{P2} -- Pairwise/2\\
\texttt{G} -- $gbest$ (fully connected) & \texttt{P3} -- Pairwise/3\\
\texttt{N} -- Von Neumann &  \\
\texttt{I} -- Increasing connectivity &  \\
\texttt{M} -- Dynamic multi-swarm &  \\
\end{tabular}
\vspace{0.5\baselineskip}
\caption{Module options and codings of velocity strategy, crossover, initialization, topology, and mutation.}
\label{tab:naming}
\end{table}
\end{small}

\section{Results} 
\label{ch:results}

\begin{figure*}%
    \includegraphics[width=.92\textwidth, right]{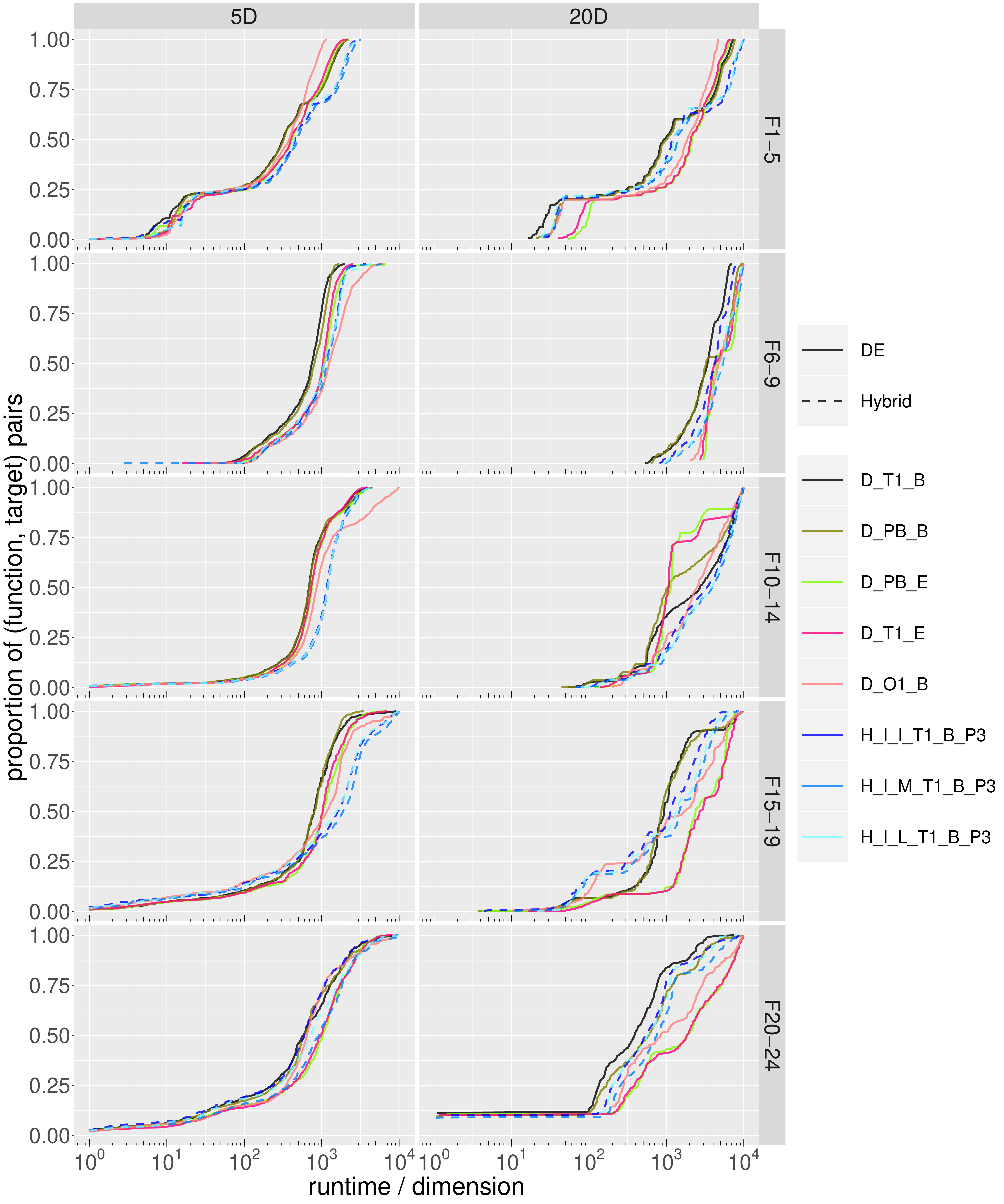}
    \caption{Empirical Cumulative Distribution Functions (ECDFs) of the top-$5$ ranked algorithms in both $5D$ and $20D$ for each function group defined in BBOB~\cite{coco}. ECDFs are aggregated over $10$ target values $10^{\{1, 0, \ldots, -8\}}$ and the ranking is in accordance with Table~\ref{tab:ERT5D} and~\ref{tab:ERT20D}. Note that only eight algorithms appear here since two algorithms are simultaneously among the top five in both $5D$ and $20D$.}%
    \label{fig:ecdf}%
\end{figure*}

Figure~\ref{fig:ecdf} depicts the Empirical Cumulative Distribution Functions (ECDF) of the top-$5$ highest ranked algorithm instances in both $5$-D and $20$-D. Due to overlap, only $8$ algorithms are shown. Tables \ref{tab:ERT5D} and \ref{tab:ERT20D} show the the Estimated Running Times of the 10 highest ranked instances, and the 10 ranked in the middle in $5$-D and $20$-D, respectively. ERT values are normalized using the corresponding ERT values of the state-of-the-art Covariance Matrix Adaptation Evolution Strategy (CMA-ES).

Though many more PSODE instances were tested, DE instances generally showed the best performance in both 5-D and 20-D. All PSO instances were outperformed by DE and many PSODE instances. This is no complete surprise, as several studies (e.g.~in \cite{psoVSde1, psoVSde2}) demonstrated the relative superiority of DE over PSO. 

Looking at the ranked algorithm instances, it is clear to see that some modules are more successful than others. The (decreasing) inertia weight velocity update strategies are dominant among the top-performing algorithms, as well as pairwise/3 selection and binomial crossover. Target-to-$p$best/1 mutation is most successful in 5-D while target-to-best/1 seems a better choice in 20-D. This is surprising, as one may expect the less greedy target-to-$p$best/1 mutation to be more beneficial in higher-dimensional search spaces, where it is increasingly difficult to avoid getting stuck in local optima. The best choice of selection method is convincingly pairwise/3. This seems to be one of the most crucial modules for the PSODE algorithm, as most instances with any other selection method show considerably worse performance. This seemingly high importance of an elitist strategy suggests that the algorithm's convergence with non-elitist selection is too slow, which could be due to the application of two different search strategies. The instances H\_I\_*\_PB\_B\_P3 and H\_I\_*\_T1\_B\_P3 appear to be the most competitive PSODE instances, with the topology choice having little influence on the observed performance. The most highly ranked DE instances are DE\_T1\_B and D\_PB\_B, in both dimensionalities. Binomial crossover seems superior to the exponential counterpart, especially in 20 dimensions.

Interestingly, the PSODE and PSO algorithms ``prefer'' different module options. As an example, the Fully Informed Particle Swarm works well on PSO instances, but PSODE instances perform better with the (decreasing) inertia weight. Bare-Bones PSO showed the overall poorest performance of the four velocity update strategies.

Notable is the large performance difference between the worst and best generated algorithm instances. Some combinations of modules, as to be expected while arbitrarily combining operators, show very poor performance, failing to solve even the most trivial problems. This stresses the importance of proper module selection.

\section{Conclusion and Future Work} \label{ch:conclusion}
We implement an extensible and modular hybridization of PSO and DE, called PSODE, in which a large number of variants from both PSO and DE is incorporated as module options. Interestingly, a vast number of unseen swarm algorithms can be easily instantiated from this hybridization, paving the way for designing and selecting appropriate swarm algorithms for specific optimization tasks.  In this work, we investigate, on $24$ benchmark functions from BBOB, $20$ PSO variants, $10$ DE variants, and $800$ PSODE instances resulting from combining the variants of PSO and DE, where we identify some promising hybrid algorithms that surpass PSO but fail to outperform the best DE variants, on subsets of BBOB problems. Moreover, we obtained insights into suitable combinations of algorithmic modules. Specifically, the efficacy of the target-to-($p$)best mutation operators, the (decreasing) inertia weight velocity update strategies, and binomial crossover was demonstrated. On the other hand, some inefficient operators, such as Bare-Bones PSO, were identified. The neighborhood topology appeared to have the least effect on the observed performance of the hybrid algorithm. 

The future work lies in extending the hybridization framework. Firstly, we are planning to incorporate the state-of-the-art PSO and DE variants as much as possible. Secondly,  we shall explore alternative ways of combining PSO and DE.  Lastly, it is worthwhile to consider the problem of selecting a suitable hybrid algorithm for an unseen optimization problem, taking the approach of automated algorithm selection. 
\input{table5D}
\input{table20D}

%% file: table5D.tex
\setlength{\tabcolsep}{0.65em}
\begin{table*}[!htbp]

\centering
\begin{tabularx}{\textwidth}{r|lrrrrrrrrr}

  \hline
 & Algorithm Instance & F1 & F2 & F6 & F8 & F11 & F12 & F17 & F18 & F21 \\ 
  \hline
 rank & CMA-ES & 658.933 & 2138.400 & 1653.667 & 2834.714 & 2207.400 & 5456.867 & 9248.600 & 13745.867 & 74140.538 \\
\hline
  1 & D\_T1\_B & 2.472 & 1.175 & 2.261 & 3.177 & 1.640 & 2.362 & 1.907 & 9.397 & 0.592 \\
  2 & D\_PB\_B & 2.546 & 1.213 & 2.321 & 4.031 & 1.643 & 2.580 & 1.258 & 5.324 & 1.072 \\
  3 & D\_PB\_E & 3.176 & 1.483 & 3.635 & 5.152 & 1.700 & 2.750 & 1.584 & 4.350 & 0.305 \\
  4 & D\_T1\_E & 3.060 & 1.477 & 3.583 & 3.670 & 1.660 & 2.281 & 2.036 & 9.112 & 0.352 \\
  5 & D\_O1\_B & 3.152 & 1.466 & 3.717 & 4.155 & 6.360 & 8.818 & 1.445 & 8.405 & 0.383 \\
  6 & H\_I\_I\_PB\_E\_P3 & 3.911 & 1.830 & 3.817 & 3.724 & 2.951 & 3.055 & 3.301 & 3.021 & 0.519 \\
  7 & H\_I\_I\_PB\_B\_P3 & 3.685 & 1.694 & 3.117 & 3.115 & 2.912 & 3.047 & 2.102 & 3.222 & 1.063 \\
  8 & H\_I\_G\_PB\_B\_P3 & 3.138 & 1.473 & 2.813 & 5.656 & 2.968 & 3.099 & 4.684 & 3.507 & 2.251 \\
  9 & H\_I\_I\_T1\_B\_P3 & 3.599 & 1.700 & 3.155 & 5.106 & 2.837 & 2.670 & 2.914 & 3.975 & 0.727 \\
  10 & H\_I\_N\_PB\_B\_P3 & 3.480 & 1.650 & 3.100 & 5.061 & 2.852 & 2.932 & 2.453 & 3.213 & 1.064 \\
  \hline 
    \dots & \dots &\dots  & \dots & \dots & \dots &\dots   & \dots & \dots & \dots &\dots \\
\hline
  411 & H\_I\_N\_PB\_B\_P2 & 4.761 & 2.268 & 4.744 & 12.933 & $\infty$ & $\infty$ & 3.113 & 53.561 & 2.738 \\
  412 & H\_D\_N\_T1\_E\_U3 & 29.656 & 38.499 & 22.459 & 25.214 & 5.091 & 9.053 & 22.333 & 8.645 & 1.247 \\
  413 & H\_B\_L\_B2\_E\_U3 & 25.515 & 13.345 & 91.998 & 10.758 & 4.203 & 5.516 & 16.277 & $\infty$ & 0.960 \\
  414 & H\_F\_L\_O1\_E\_U3 & 19.585 & 9.980 & 94.563 & 18.771 & 5.529 & 12.265 & 161.662 & 7.416 & 3.586 \\
  415 & H\_B\_G\_T1\_E\_P3 & 4.736 & 2.288 & 6.532 & 10.503 & $\infty$ & 45.093 & 2.808 & 36.108 & 3.474 \\
  416 & H\_B\_N\_B1\_B\_U2 & 6.531 & 3.029 & 8.313 & 6.918 & 93.749 & 13.117 & 28.817 & $\infty$ & 19.629 \\
  417 & H\_D\_I\_T1\_E\_P2 & 5.506 & 2.545 & 5.917 & 12.812 & $\infty$ & $\infty$ & 7.791 & 34.691 & 3.433 \\
  418 & H\_D\_M\_O1\_E\_U3 & 21.270 & 10.963 & 33.571 & 12.992 & 5.882 & 7.250 & 12.577 & 5.760 & 1.192 \\
  419 & H\_B\_G\_O1\_B\_P2 & 4.091 & 1.764 & 4.959 & $\infty$ & $\infty$ & $\infty$ & 157.845 & $\infty$ & 2.253 \\
  420 & H\_F\_L\_T1\_E\_U3 & 26.450 & 15.383 & 17.706 & 12.174 & 4.609 & 9.334 & 16.892 & 53.541 & 1.822 \\
\end{tabularx}
\caption{On $5D$, the normalized Expected Running Time (ERT) values of the top-$10$ ranked and $10$ algorithms ranked in the middle among all $830$ algorithms. The ranking is firstly determining on each test problem with respect to ERT and then averaged over all $24$ test problem. For the reported ERT values, the target $f_{\text{opt}} + 10^{-7}$ is used. All ERT values are normalized per problem with respect to a reference CMA-ES, shown in the first row of algorithms.} \label{tab:ERT5D}
\vspace{-0.4cm}
\end{table*}

%% file: table20D.tex
\begin{table*}[!htbp]
\centering
\scalebox{1}{\begin{tabularx}{\textwidth}{r|lrrrrrrrrrrrr}

  \hline
 & Algorithm Instance & F1 & F2 & F6 & F8 & F11 & F12 & F17 & F18 & F21 \\ 
  \hline
rank & CMA-ES & 830.800 & 16498.533 & 4018.600 & 19140.467 & 12212.267 & 15316.733 & 5846.400 & 17472.333 & 801759 \\ 
\hline
  1 & D\_T1\_B & 7.377 & 0.864 & 5.912 & 3.702 & 2.678 & 4.699 & 3.144 & 3.604 & 0.385 \\ 
  2 & D\_PB\_B & 7.731 & 0.901 & 6.884 & 6.766 & 3.833 & 5.999 & 3.158 & 1.719 & 0.193 \\ 
  3 & H\_I\_I\_T1\_B\_P3 & 10.988 & 1.195 & 7.894 & 4.153 & 6.596 & 7.656 & 3.988 & 3.081 & 0.298 \\ 
  4 & H\_I\_M\_T1\_B\_P3 & 12.621 & 1.434 & 9.714 & 5.296 & 8.389 & 8.152 & 4.979 & 3.138 & 0.186 \\ 
  5 & H\_I\_L\_T1\_B\_P3 & 11.402 & 1.299 & 9.271 & 5.146 & 8.170 & 7.422 & 4.771 & 3.406 & 0.341 \\ 
  6 & H\_I\_N\_T1\_B\_P3 & 10.641 & 1.202 & 8.218 & 4.705 & 7.253 & 7.928 & 4.325 & 2.741 & 0.338 \\ 
  7 & H\_D\_M\_T1\_B\_P3 & 12.865 & 1.476 & 10.100 & 6.036 & 8.119 & 8.768 & 5.345 & 3.450 & 0.354 \\ 
  8 & D\_B2\_B & 7.983 & 0.885 & $\infty$ & 5.862 & 10.401 & 6.455 & 9.258 & 44.240 & 0.829 \\ 
  9 & H\_D\_G\_T1\_B\_P3 & 9.031 & 1.074 & 7.910 & 4.419 & 5.690 & 8.078 & 4.079 & 7.838 & 0.695 \\ 
  
  10 & H\_D\_N\_T1\_B\_P3 & 11.307 & 1.287 & 9.057 & 4.801 & 9.854 & 5.949 & 4.517 & 4.288 & 0.303 \\ 
    \hline 
    \dots & \dots &\dots  & \dots & \dots & \dots &\dots   & \dots & \dots & \dots &\dots \\
\hline
  411 & H\_D\_L\_T1\_B\_U2 & 39.225 & 6.262 & $\infty$ & 312.925 & $\infty$ & 35.178 & $\infty$ & $\infty$ & 0.728 \\ 
  412 & H\_F\_M\_T1\_B\_U2 & 55.045 & 5.655 & $\infty$ & $\infty$ & $\infty$ & 34.213 & $\infty$ & $\infty$ & 0.360 \\ 
  413 & H\_B\_M\_T1\_E\_P2 & 39.181 & 4.393 & $\infty$ & $\infty$ & $\infty$ & 41.771 & $\infty$ & $\infty$ & 0.369 \\ 
  414 & P\_F\_N & 53.733 & $\infty$ & 1480.838 & $\infty$ & 88.421 & $\infty$ & $\infty$ & $\infty$ & 0.163 \\ 
  415 & H\_I\_M\_T1\_B\_U2 & 40.014 & 7.379 & $\infty$ & 313.468 & $\infty$ & 35.252 & $\infty$ & $\infty$ & 0.546 \\ 
  416 & H\_I\_N\_PB\_B\_U3 & 70.776 & 362.611 & 86.426 & $\infty$ & 18.979 & $\infty$ & 339.045 & 113.442 & 0.433 \\ 
  417 & H\_I\_M\_B1\_E\_P2 & 33.073 & 3.734 & $\infty$ & $\infty$ & $\infty$ & 72.629 & 35.327 & $\infty$ & 0.876 \\ 
  418 & H\_I\_G\_B2\_B\_U2 & 43.424 & 8.498 & $\infty$ & 104.122 & $\infty$ & $\infty$ & $\infty$ & $\infty$ & 7.367 \\ 
  419 & H\_B\_G\_PB\_B\_U2 & 41.308 & 16.007 & $\infty$ & $\infty$ & $\infty$ & 50.786 & $\infty$ & $\infty$ & 1.054 \\ 
  420 & H\_B\_N\_PB\_B\_P3 & 33.984 & 4.203 & $\infty$ & $\infty$ & $\infty$ & 32.929 & $\infty$ & $\infty$ & 1.314 \\ 
   \hline
\end{tabularx}}
\centering
\centering
\caption{On $20D$, the normalized Expected Running Time (ERT) values of the top-$10$ ranked and $10$ algorithms ranked in the middle among all $830$ algorithms. The ranking is firstly determining on each test problem with respect to ERT and then averaged over all $24$ test problem. For the reported ERT values, the target $f_{\text{opt}} + 10^{-1}$ is used. All ERT values are normalized per problem with respect to a reference CMA-ES, shown in the first row of algorithms.}\ \label{tab:ERT20D}
\end{table*}